# AutoFormBench: Benchmark Dataset for Automating Form Understanding

Gaurab Baral[1], Junxiu Zhou[2]
School of Computing and Analytics, Northern Kentucky University
Highland Heights, Kentucky, 41076
baralg1@nku.edu[1], zhouj2@nku.edu[2]

**ABSTRACT**

*Automated processing of structured documents such as government forms, healthcare records, and enterprise invoices remains a persistent challenge due to the high degree of layout variability encountered in real-world settings. This paper introduces AutoFormBench, a benchmark dataset of 407 annotated real-world forms spanning government, healthcare, and enterprise domains, designed to train and evaluate form element detection models. We present a systematic comparison of classical OpenCV approaches and four YOLO architectures (YOLOv8, YOLOv11, YOLOv26-s, and YOLOv26-l) for localizing and classifying fillable form elements. specifically checkboxes, input lines, and text boxes across diverse PDF document types. YOLOv11 demonstrates consistently superior performance in both F1 score and Jaccard accuracy across all element classes and tolerance levels.*

## I. INTRODUCTION

The digitization of administrative workflows has created a growing demand for systems that can automatically interpret and process structured documents. Forms such as government applications, healthcare records, and enterprise invoices are among the most common document types encountered in real-world pipelines [1]. Yet despite widespread digital adoption, many of these documents still exist as scanned images or fixed-layout PDFs, making automated field extraction a persistent challenge [2].

Traditional approaches rely on either manual data entry or rigid template-matching systems that assume a fixed document structure. These methods fail quickly when faced with real-world variability such as skewed scans, inconsistent formatting, and diverse layouts across domains [3]. Recent advances in object detection, particularly the YOLO (You Only Look Once) family of models, have shown strong potential for localizing document elements at practical inference speeds [4]. However, detection alone is not sufficient. A complete pipeline must identify where a field exists, classify its functional type, and map that spatial information into a structured, machine-readable format [5].

This paper presents AutoFormBench, a benchmark dataset of 407 annotated real-world forms spanning government, healthcare, and enterprise domains, constructed to support the training and evaluation of form element detection models. We conduct a systematic comparison of classical OpenCV approaches and multiple YOLO architectures for localizing and classifying fillable form elements: checkboxes, input lines, and text boxes across diverse PDF document types [6].

## II. RELATED WORK

### A. Classical Computer Vision Approaches

The evolution of Document Layout Analysis (DLA) has its roots in deterministic, rule-based computer vision methods. Breuel (2003) laid foundational groundwork in this area by introducing geometric algorithms for identifying maximal whitespace rectangles and recovering text line structures through constrained maximum-likelihood matching, enabling layout analysis without the need for heuristic tuning [7]. While computationally efficient, such rule-based methods are inherently brittle in the presence of real-world document noise including skewed scans, overlapping text, and varying border thicknesses, limitations directly encountered in the primary dataset of government, health, and enterprise forms used in this work.

### B. Deep Learning for Document Layout Analysis

With the advent of deep learning, the treatment of document elements as objects to be detected transformed the field. Zhong et al. (2019) introduced PubLayNet, the largest document layout dataset at the time of its release, containing over 360,000 annotated document images sourced from PubMed Central [8]. PubLayNet demonstrated that architectures originally designed for natural scene object detection specifically Faster-RCNN and Mask-RCNN could be effectively repurposed for identifying document regions such as text, titles, tables, figures, and lists, achieving macro average mAP scores exceeding 0.9. This established transfer learning from vision datasets as a viable paradigm for document intelligence. More recently, the YOLO family of models has gained traction in the document analysis community owing to its superior inference speed and competitive localization precision [4, 9]. Research by Deng et al. (2024) further explored YOLO-based architectures for DLA, incorporating multi-convolutional deformable separation (MCDS) modules specifically designed to handle the high variability in document element aspect ratios [10].

### C. Specialized Datasets for Form Field Detection

As document analysis matured, research attention shifted toward datasets designed specifically for structured form understanding. Jaume et al. (2019) introduced FUNSD (Form Understanding in Noisy Scanned Documents), the first publicly available benchmark with comprehensive annotations for form understanding tasks [11]. Comprising 199 fully annotated scanned forms, FUNSD supports a range of tasks including text detection, spatial layout analysis, and entity labeling, and established a standardized evaluation protocol that has since become widely adopted in the field. More recently, the introduction of CommonForms represents the first web-scale dataset specifically for form field detection. By filtering 8 million PDFs from Common Crawl into a curated set of 450,000 pages, CommonForms enables the training of specialized models like FFDNet, demonstrating that high-resolution inputs and diverse, multi-language training data are essential for achieving the precision necessary for modern IDP pipelines [12].

## III. DATASET DESCRIPTIONS

To evaluate the proposed detection framework, a complete experimental environment was established. This section describes the collection and preparation of the custom form dataset, the annotation and labeling protocol, and the training procedures used for each detection approach.

### A. Data Collection

The primary challenge in automated form filling is the high degree of structural variability across different document providers. To address this, a diverse dataset of approximately 506 PDF documents was curated, focusing on three primary categories: government forms, health and insurance forms, and enterprise invoices. All documents were converted into high-resolution images at 300 DPI to ensure that fine-grained elements, such as radio buttons and small-print labels, retained sufficient visual features for the detection pipeline. This corpus provides a balanced representation of real-world document noise, including varying font weights, overlapping lines, and diverse box-border thicknesses. The entire dataset can be viewed at https://tempgaurab.github.io/Detect_Regions_in_PDF/#download .Few sample forms are shown below in Figure 1.

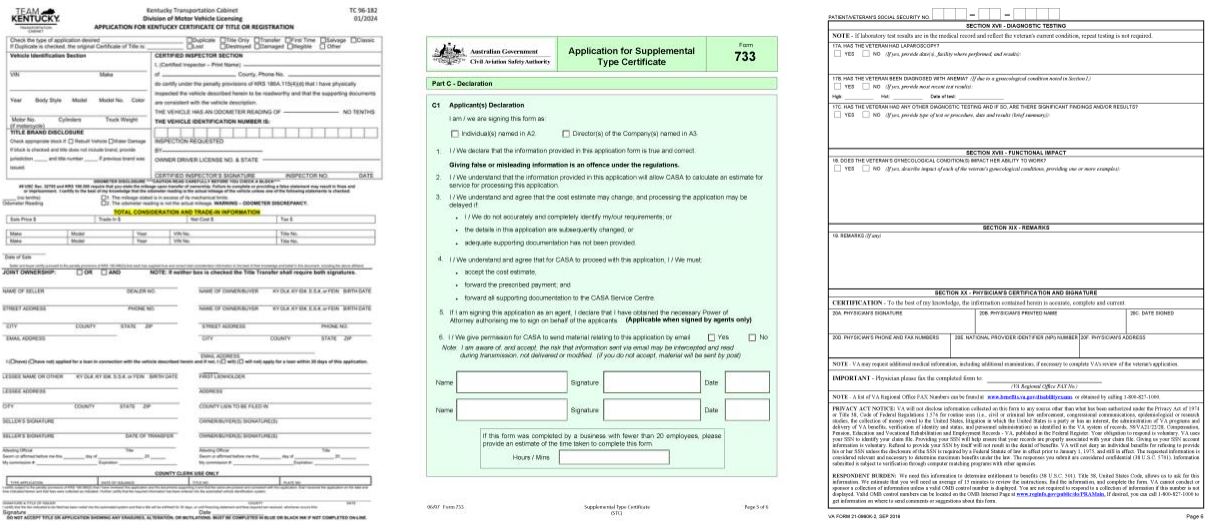

*[Figure 1: 3 sample forms used in this analysis extracted from several sources in the internet.]*

*B. Annotation and Labeling*

To establish a reliable ground truth for model training and evaluation, all 506 collected PDF documents underwent a structured manual annotation process. Prior to labeling, 99 documents were identified as empty; containing no detectable fillable elements and were subsequently excluded from the corpus, yielding a final annotated set of 407 documents.

Annotation was performed using a custom-built Python application developed specifically for this workflow. The tool rendered each PDF page as a high-resolution image and provided an interactive interface through which annotators could manually define and classify regions of interest. Each annotated region was assigned to one of three predefined categories: checkboxes, lines, and boxes, corresponding to the primary structural elements observed across the document corpus. For each document, the bounding coordinates of every labeled region were recorded and exported as structured JSON files. These JSON files served as the ground truth reference against which all subsequent detection models were evaluated. A sample json file is shown below:

```
{
  "checkboxes": [
    [537.5,  478.12],
    [759.38, 478.12],
    [987.5,  478.12]
  ],
  "lines": [
    [262.5,  1365.62, 612.5,   1368.75],
    [300.0,  1440.62, 609.38,  1434.38],
    [1600.0, 1575.0,  1062.5,  1581.25],
    [1056.25,1653.12, 1596.88, 1650.0 ],
    [968.75, 1753.12, 1012.5,  1753.12],
    [1137.5, 1753.12, 1453.12, 1753.12],
    [1506.25,1756.25, 1581.25, 1756.25],
    [1081.25,1893.75, 1600.0,  1893.75]
  ],
  "boxes": [
    [131.25, 306.25, 478.12, 365.6]
}
```

## IV. EXPERIMENTS

The experimental setup encompassed two distinct detection paradigms: a classical computer vision approach using OpenCV, and a deep learning approach using a family of YOLO-based object detection architectures.

### A. OpenCV-Based Detection

The OpenCV pipeline requires no training, operating entirely through deterministic geometric analysis. Images are first converted to 8-bit grayscale and binarized using Otsu's thresholding, with a light morphological closing applied to repair broken contour edges. Contours are extracted using findContours and filtered to retain only four-vertex quadrilaterals via Douglas-Peucker approximation. Each candidate region is then classified geometrically: near-square regions within 10–50 pixels are labeled as checkboxes, while wide shallow rectangles with an aspect ratio exceeding 1.5 are classified as text input fields. Full-page bounding boxes are explicitly discarded. Two configurations were evaluated: an Initial variant using baseline geometric thresholds, and an Advanced variant incorporating additional filtering heuristics and stricter contour validation to reduce false positives.

### B. YOLO-Based Detection

Four YOLO architectures: YOLOv8, YOLOv11, YOLOv26-s, and YOLOv26-l were each trained on the annotated dataset of 407 documents using a consistent protocol to ensure fair and reproducible comparison. Annotated JSON ground truth files were converted into YOLO-compatible label format, representing each bounding box as a normalized center coordinate, width, and height tuple paired with its class index. The dataset was partitioned into training, validation, and test splits following a 70/15/15 ratio, with stratification applied to maintain proportional class representation across all subsets.

Each model was initialized with pretrained weights from the COCO benchmark. Data augmentation techniques applied during training included random horizontal flipping, mosaic composition, and scale jittering. Input images were resized to 640×640 pixels. All models were trained for a maximum of 100 epochs with early stopping based on validation mAP, using the Adam optimizer with an initial learning rate of 0.001 and a cosine annealing decay schedule. Batch size was set to 16 across all experiments, and model checkpoints were saved at the epoch of highest validation performance.

## V. RESULTS AND EVALUATION

### A. Evaluation Methodology

All detection models were evaluated against the manually annotated ground truth using a tolerance-based matching protocol. A predicted bounding box is considered a true positive if each of its four edge coordinates lies within a specified tolerance of the corresponding ground truth coordinate. Three tolerance levels were tested: 5%, 10%, and 20% of the image dimensions. This design reflects the practical reality of form-filling automation, where minor coordinate offsets are acceptable as long as the field is correctly localized. Performance is reported using four metrics per class: Precision, Recall, F1 Score, and Jaccard Accuracy (Intersection over Union), computed globally by accumulating true positives, false positives, and false negatives across all 407 test documents.

### B. Classical OpenCV Results

Table I presents the results for both OpenCV configurations across all three tolerance levels. The Initial OpenCV pipeline demonstrated strong checkbox precision (0.817–0.822) but performed poorly on lines (F1: 0.291–0.341) and boxes (F1: 0.374–0.474), reflecting the fundamental difficulty of distinguishing fine-grained horizontal strokes and variable-border boxes using only contour geometry. The Advanced OpenCV variant, which introduced stricter contour validation and label-proximity filtering for lines, improved line detection substantially (F1: 0.418–0.454) and box recall at the 20% tolerance level (0.724). However, this came at the cost of checkbox precision, which dropped from 0.822 to 0.685 at 20% tolerance, as the more aggressive filtering occasionally rejected valid checkbox contours. Neither OpenCV variant achieved competitive performance on lines and boxes relative to the deep learning approaches, confirming the limitations of purely geometric methods on structurally variable documents.

TABLE I. OpenCV Detection Performance

*Tolerance: 5%*

| Model | Class | Precision | Recall | F1 | Jaccard |
|---|---|---|---|---|---|
| Initial OpenCV | Checkboxes | 0.817 | 0.683 | 0.744 | 0.592 |
|  | Lines | 0.215 | 0.453 | 0.291 | 0.170 |
|  | Boxes | 0.273 | 0.590 | 0.374 | 0.230 |
| Advanced OpenCV | Checkboxes | 0.679 | 0.707 | 0.693 | 0.530 |
|  | Lines | 0.515 | 0.352 | 0.418 | 0.264 |
|  | Boxes | 0.320 | 0.571 | 0.410 | 0.258 |

*Tolerance: 10%*

| Model | Class | Precision | Recall | F1 | Jaccard |
|---|---|---|---|---|---|
| Initial OpenCV | Checkboxes | 0.819 | 0.685 | 0.746 | 0.595 |
|  | Lines | 0.228 | 0.480 | 0.309 | 0.183 |
|  | Boxes | 0.295 | 0.638 | 0.404 | 0.253 |
| Advanced OpenCV | Checkboxes | 0.682 | 0.710 | 0.696 | 0.533 |
|  | Lines | 0.522 | 0.356 | 0.423 | 0.268 |
|  | Boxes | 0.345 | 0.615 | 0.442 | 0.284 |

*Tolerance: 20%*

| Model | Class | Precision | Recall | F1 | Jaccard |
|---|---|---|---|---|---|
| Initial OpenCV | Checkboxes | 0.822 | 0.687 | 0.749 | 0.598 |
| | Lines | 0.251 | 0.530 | 0.341 | 0.205 |
| | Boxes | 0.347 | 0.750 | 0.474 | 0.311 |
| Advanced OpenCV | Checkboxes | 0.685 | 0.713 | 0.699 | 0.537 |
| | Lines | 0.559 | 0.382 | 0.454 | 0.294 |
| | Boxes | 0.406 | 0.724 | 0.520 | 0.351 |

### C. YOLO Architecture Comparison

Table II compares the four YOLO architectures evaluated in this study. Across all tolerance levels and element classes, YOLOv11 demonstrated the strongest and most consistent overall performance. At the 10% tolerance level, the most operationally relevant threshold, YOLOv11 achieved F1 scores of 0.817, 0.815, and 0.658 for checkboxes, lines, and boxes respectively, outperforming YOLOv8 on all three classes and exceeding both YOLOv26 variants on the critical lines and boxes classes.

The YOLOv26 variants exhibited a characteristic precision-recall imbalance, particularly pronounced in the large variant (YOLOv26-l). YOLOv26-l achieved the highest checkbox precision of any model (0.981 at 20% tolerance) but at a severe recall cost (0.575), yielding an F1 of only 0.725, notably below YOLOv11's 0.827. This behavior suggests that the larger YOLOv26 model is over-conservative in its predictions on the custom document dataset, likely due to the domain gap between its pretraining distribution and the form-specific visual characteristics encountered here. YOLOv8, while competitive with YOLOv11 on recall, consistently lagged on precision across all three classes, particularly for boxes at stricter tolerance levels.

**TABLE II. YOLO Architecture Comparison**

*Tolerance: 5%*

| Model | Class | Precision | Recall | F1 | Jaccard |
|---|---|---|---|---|---|
| YOLOv8 | Checkboxes | 0.789 | 0.794 | 0.791 | 0.655 |
| | Lines | 0.715 | 0.824 | 0.766 | 0.620 |
| | Boxes | 0.540 | 0.642 | 0.587 | 0.415 |
| **YOLOv11** | **Checkboxes** | **0.827** | **0.774** | **0.800** | **0.666** |
| | **Lines** | **0.752** | **0.829** | **0.789** | **0.652** |
| | **Boxes** | **0.587** | **0.645** | **0.615** | **0.444** |
| YOLOv26-s | Checkboxes | 0.912 | 0.601 | 0.725 | 0.569 |
| | Lines | 0.793 | 0.697 | 0.742 | 0.590 |
| | Boxes | 0.571 | 0.588 | 0.579 | 0.408 |
| YOLOv26-l | Checkboxes | 0.958 | 0.562 | 0.708 | 0.549 |
| | Lines | 0.790 | 0.784 | 0.787 | 0.649 |
| | Boxes | 0.624 | 0.634 | 0.629 | 0.459 |

*Tolerance: 10%*

| Model | Class | Precision | Recall | F1 | Jaccard |
|---|---|---|---|---|---|
| YOLOv8 | Checkboxes | 0.808 | 0.813 | 0.811 | 0.682 |

| Model | Class | Precision | Recall | F1 | Jaccard |
|---|---|---|---|---|---|
| | Lines | 0.736 | 0.847 | 0.788 | 0.650 |
| | Boxes | 0.574 | 0.682 | 0.623 | 0.453 |
| YOLOv11 | Checkboxes | 0.845 | 0.790 | 0.817 | 0.690 |
| | **Lines** | **0.777** | **0.857** | **0.815** | **0.688** |
| | **Boxes** | **0.628** | **0.690** | **0.658** | **0.490** |
| YOLOv26-s | Checkboxes | 0.933 | 0.615 | 0.741 | 0.589 |
| | Lines | 0.817 | 0.717 | 0.764 | 0.618 |
| | Boxes | 0.615 | 0.633 | 0.624 | 0.454 |
| YOLOv26-l | Checkboxes | 0.973 | 0.571 | 0.719 | 0.562 |
| | Lines | 0.806 | 0.799 | 0.802 | 0.670 |
| | Boxes | 0.663 | 0.673 | 0.668 | 0.502 |

*Tolerance: 20%*

| Model | Class | Precision | Recall | F1 | Jaccard |
|---|---|---|---|---|---|
| YOLOv8 | Checkboxes | 0.820 | 0.826 | 0.823 | 0.699 |
| | Lines | 0.755 | 0.869 | 0.808 | 0.677 |
| | Boxes | 0.667 | 0.793 | 0.724 | 0.568 |
| **YOLOv11** | **Checkboxes** | **0.855** | **0.800** | **0.827** | **0.705** |
| | **Lines** | **0.797** | **0.878** | **0.836** | **0.718** |
| | **Boxes** | **0.726** | **0.798** | **0.760** | **0.613** |
| YOLOv26-s | Checkboxes | 0.944 | 0.622 | 0.750 | 0.600 |
| | Lines | 0.839 | 0.737 | 0.785 | 0.646 |
| | Boxes | 0.713 | 0.734 | 0.724 | 0.567 |
| YOLOv26-l | Checkboxes | 0.981 | 0.575 | 0.725 | 0.569 |
| | Lines | 0.825 | 0.818 | 0.821 | 0.697 |
| | Boxes | 0.761 | 0.773 | 0.767 | 0.622 |

Based on these results, YOLOv11 demonstrates superior performance in both F1 score and Jaccard accuracy, which may be attributed to the fine-tuning performed on a relatively small dataset. The training dynamics of the YOLOv11 model were analyzed using the output artifacts generated during the training run. Figure 2 presents the training and validation loss curves alongside precision, recall, and mAP metrics over the full 100-epoch training cycle. All three loss components — box loss, classification loss, and distribution focal loss (DFL) — exhibit rapid initial descent within the first 20 epochs followed by stable convergence, with training and validation curves tracking closely and showing no significant divergence indicative of overfitting. Precision and recall on the validation set rise steadily, reaching approximately 0.65 by epoch 100. The mAP@0.5 curve plateaus near 0.60, while mAP@0.50:0.95 stabilizes around 0.30, consistent with the expected performance range for a single-stage detector on a custom domain-specific dataset with fine-grained element classes.

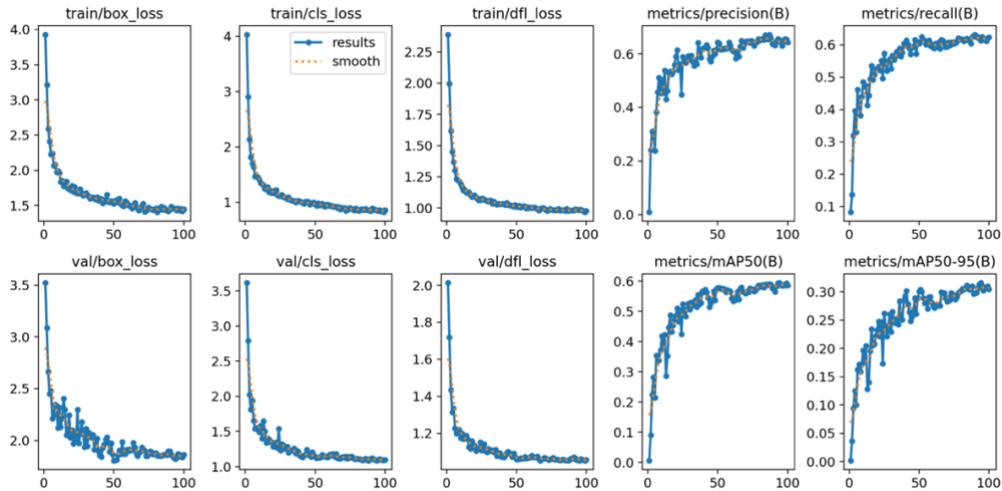

*[Figure 2: YOLOv11 training and validation loss curves (box loss, cls loss, DFL loss) and validation metrics (precision, recall, mAP@0.5, mAP@0.50:0.95) over 100 training epochs.]*

Figure 3 presents the normalized confusion matrix, which provides per-class breakdown of detection accuracy. The diagonal entries reveal a clear performance gradient across element types: boxes achieve the highest correct classification rate (0.80), followed by lines (0.66), while checkboxes show the lowest true positive rate (0.49). The primary source of error across all three classes is suppression rather than misclassification, 51% of true checkboxes, 34% of true lines, and 20% of true boxes are absorbed into the background, indicating that the model errs on the side of under-predicting rather than producing spurious detections. The background row, however, tells a more nuanced story: a substantial proportion of true background regions are incorrectly predicted as checkboxes (0.73), with smaller fractions misclassified as lines (0.12) and boxes (0.15). This asymmetry suggests that the model's checkbox detector is sensitive to background textures, such as printed tick marks, stamps, or bordered cell corners that share visual features with actual checkboxes. Taken together, the matrix indicates that the dominant failure mode is missed detection for lines and boxes, while for checkboxes it is a combination of missed detection and elevated false positives on visually ambiguous background regions, both of which are consistent with the relatively modest dataset size and the fine-grained visual similarity between form elements and non-field document structures.

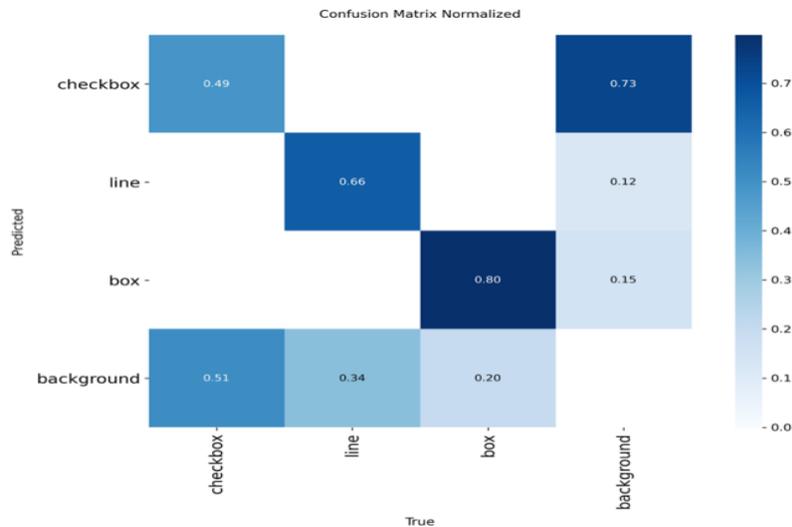

*[Figure 3: Normalized confusion matrix for YOLOv11 across the three element classes (Checkboxes, Lines, Boxes) and background.]*

## VI. DISCUSSION

The results demonstrate a clear performance hierarchy across the two detection paradigms evaluated. Classical OpenCV approaches, while effective for checkboxes where structural regularity is high, proved fundamentally insufficient for lines and boxes, the Initial OpenCV variant's line F1 of 0.291 reflects an inherent limitation of contour-based methods rather than a tuning failure, as horizontal strokes share geometric properties with decorative borders and table dividers that cannot be disambiguated without learned context. The counter-intuitive underperformance of the larger YOLOv26 variants relative to YOLOv11 is equally notable: YOLOv26-l's checkbox precision of 0.981 is the highest recorded by any model, yet its recall of 0.575 collapses the F1 to 0.725. This pattern suggests a domain adaptation problem, larger models pretrained on natural scene datasets carry stronger priors that resist full adaptation to a relatively small 407-document form corpus, while YOLOv11's more modest pretraining footprint allows faster convergence to the form detection domain, producing the balanced precision-recall profile that yields the highest standalone F1 and Jaccard scores.

The consistent metric improvement as tolerance increases from 5% to 20% across all models indicates that many predictions are geometrically correct in spatial localization but offset by small pixel-level distances, a pattern characteristic of rendering variation across PDF sources rather than fundamental misdetection. For downstream form-filling applications, where the goal is field region localization rather than exact coordinate reconstruction, the 10–20% tolerance regime is operationally relevant. The mAP@0.5 plateau near 0.60 observed in the YOLOv11 training curves suggests that further gains remain available through dataset expansion, and the manually annotated 407-document corpus established here provides a quality-controlled foundation for that work.

## VII. FUTURE WORK

Several promising directions remain for advancing this research on AI-assisted form filling. First, extending the current approach to digital and semi-structured forms could be explored. By developing specialized prompting strategies for large language models or dedicated AI agents, the system could dynamically understand document context, infer missing information, and automatically populate fields — building on emerging techniques in intelligent document understanding and automated data extraction.

Second, expanding the training dataset through additional high-quality annotations from diverse annotators would likely improve model robustness and generalization. Incorporating more varied form layouts, languages, and domains could reduce detection errors in edge cases and enhance overall accuracy across underrepresented document types. Finally, incorporating mechanisms for uncertainty estimation and abstention represents an important next step. This could be achieved through calibration methods, ensemble models, or explicit reject options during inference, thereby increasing reliability in real-world deployment scenarios where erroneous predictions carry high costs.